\documentclass[10pt,twocolumn,letterpaper]{article}

\usepackage{caption}
\usepackage{style/iccv}
\usepackage{times}
\usepackage{epsfig}
\usepackage{graphbox} %
\usepackage[export]{adjustbox}
\usepackage{amsmath}
\usepackage{amssymb}
\usepackage{float}
\usepackage{color, colortbl}
\usepackage{booktabs}
\usepackage{authblk}

\usepackage[pagebackref=true,breaklinks=true,letterpaper=true,colorlinks,bookmarks=false]{hyperref}

\iccvfinalcopy %

\def\iccvPaperID{6467} %

\ificcvfinal\fi
\begin{document}

\newcommand{\TODO}[1]{}
\renewcommand{\TODO}[1]{{\color{red} TODO: {#1}}}
\title{Learning Shape Templates with Structured Implicit Functions}

\makeatletter
\renewcommand\Authfont{\fontsize{11.5}{14.4}\selectfont}
\renewcommand\AB@affilsepx{\qquad \protect\Affilfont}
\makeatother
\author[1,2]{Kyle Genova}
\author[2]{Forrester Cole}
\author[2]{Daniel Vlasic}
\author[2]{Aaron Sarna}
\author[2]{William T. Freeman}
\author[1, 2]{Thomas Funkhouser}
\affil[1]{Princeton University}
\affil[2]{Google Research}
\renewcommand*{\Authsep}{ }%
\renewcommand*{\Authands}{ }%

\twocolumn[{%
\renewcommand\twocolumn[1][]{#1}%
\maketitle
\begin{center}
    \centering
    \vspace{-1.5em}
    \includegraphics[width=\textwidth]{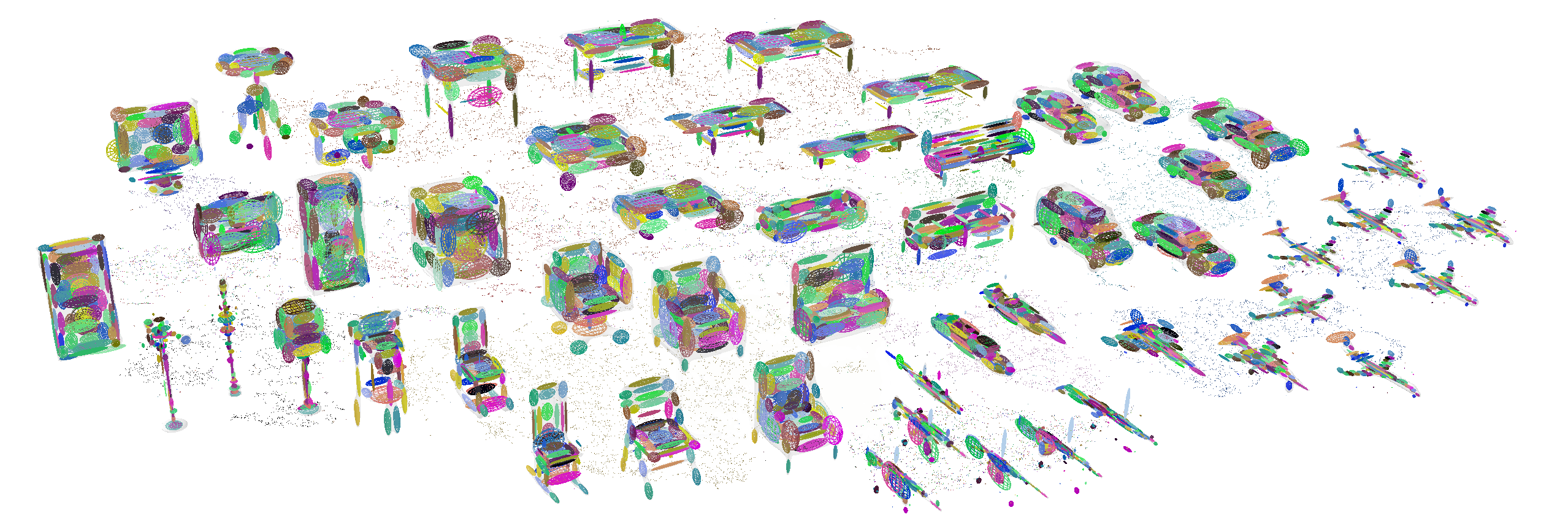}
    \captionof{figure}{Shapes from the ShapeNet~\cite{chang2015shapenet} database, fit to a structured implicit template, and arranged by template parameters using t-SNE~\cite{vanDerMaaten2008tsne}. Similar shape classes, such as airplanes, cars, and chairs, naturally cluster by template parameters.\protect\footnotemark[1]}
    \label{fig:3d_tsne}
\end{center}%
}]

\begin{abstract}
\vspace{-0.75em}
Template 3D shapes are useful for many tasks in graphics and vision, including fitting observation data, analyzing shape collections, and transferring shape attributes. Because of the variety of geometry and topology of real-world shapes, previous methods generally use a library of hand-made templates. In this paper, we investigate learning a general shape template from data. To allow for widely varying geometry and topology, we choose an implicit surface representation based on composition of local shape elements. While long known to computer graphics, this representation has not yet been explored in the context of machine learning for vision. We show that \emph{structured implicit} functions are suitable for learning and allow a network to smoothly and simultaneously fit multiple classes of shapes. The learned shape template supports applications such as shape exploration, correspondence, abstraction, interpolation, and semantic segmentation from an RGB image.

\end{abstract}

\section{Introduction}

\begin{figure}[t]
    \centering
    \includegraphics[width=\columnwidth]{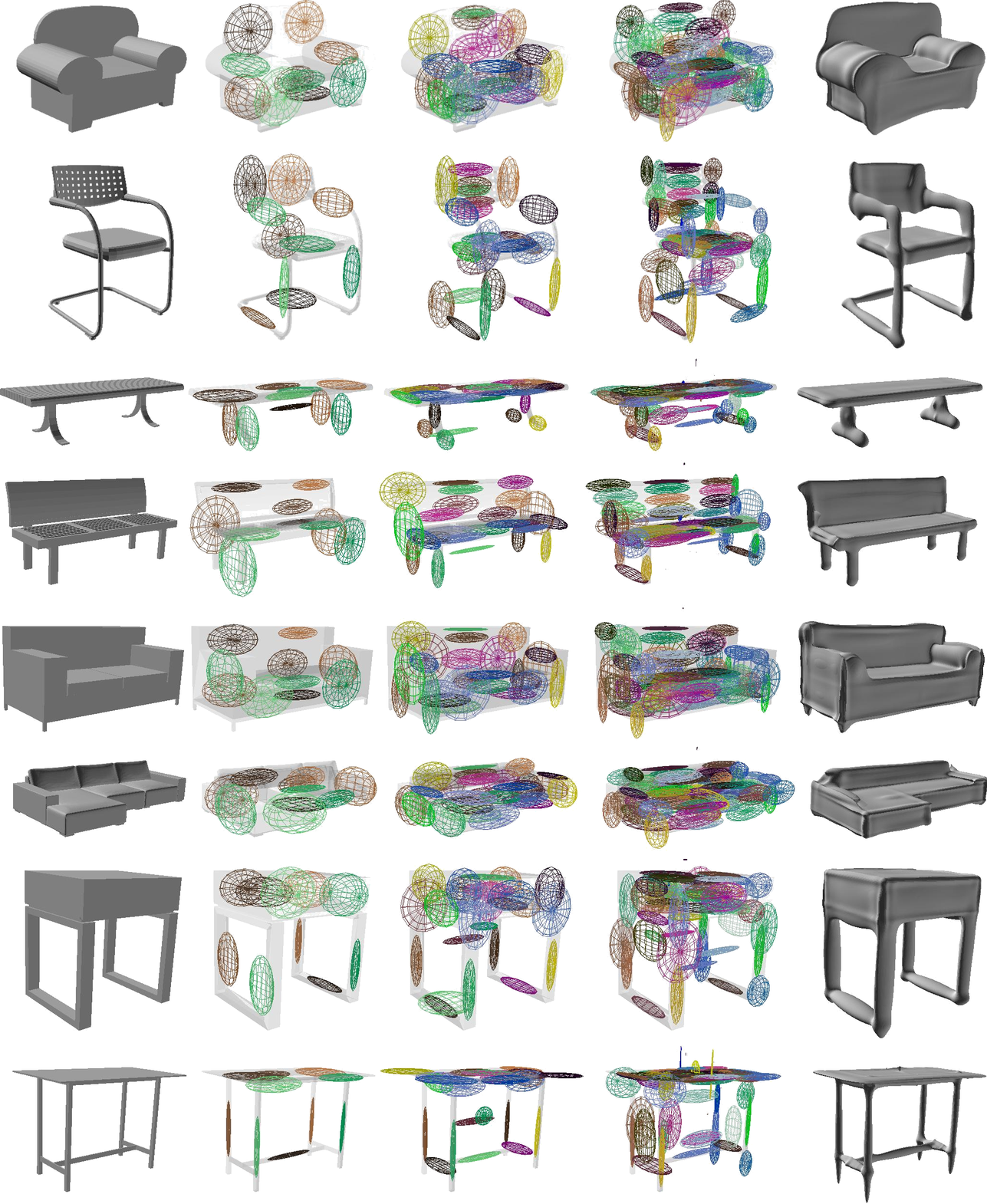}
    \caption{Templates fit to a variety of geometry and topology. Middle columns: three shape templates trained across classes with 10, 25, and 100 elements, respectively. Right: surface reconstruction of the implicit function defined by the 100 element template. 
    Note how the structure of each template is consistent between shapes.
    }
    \label{fig:multiclass_examples}
\end{figure}

Fitting a 3D shape template to observations is one of the oldest and most durable vision techniques~\cite{roberts1963}. Templates offer a concise representation of complex shapes and a strong prior for fitting. They can be used to directly correspond and compare shapes, and supervised learning approaches may be applied to correspond the template and a photograph~\cite{wei2016dense,bogo2016smplify}. In order to fit a wide range of shapes, however, multiple, hand-made templates are usually required, along with a procedure for choosing the appropriate one~\cite{ganapathi2018parsing}.

The goal of this paper is to construct a general shape template that fits any shape, and to learn the parameters of this template from data. \footnotetext[1]{See \url{templates.cs.princeton.edu} for video, supplemental, and high resolution results.} We view a shape as a level set of a volumetric function and approximate that function by a collection of shape elements with local influence, a formulation we term a \emph{structured implicit} function.
The template itself is defined by the number of and formula for the shape elements, and the template parameters are simply the concatenation of the parameters of each element. An example of this type of representation is the classic \emph{metaballs} method~\cite{blinn1982generalization}, but more sophisticated versions have been proposed since~\cite{wyvill1986soft,bloomenthal1991convolution,ohtake2003multi}.

Given a template definition, we show that a network can be trained to fit the template to shapes with widely varying geometry and topology (Figure~\ref{fig:3d_tsne}). Critically, the network learns a fitting function that is smooth: the template parameters of similar shapes are similar, and vary gradually through shape-space (Figure~\ref{fig:multiclass_examples}). Further, we show that the network learns to associate each shape element with similar structures in each shape: for example, the tail fin of an aircraft may be represented by one element, while the left wingtip may be represented by another. This consistency allows us to interpolate shapes, estimate vertex correspondences, or predict the influence region of a given element in a 2D image, providing semantic segmentation of shapes.

The closest related work to ours is the volumetric primitive approach of Tulsiani, et al.~\cite{tulsiani2017learning}. Like that work, we aim to learn a consistent shape representation with a small number of primitives. We expand on their work by specifying the surface as a structured implicit function, rather than as a collection of explicit surface primitives. This change allows for an order of magnitude increase in the number of shape elements, allowing our template to capture fine details.

Our method is entirely self-supervised and requires only a collection of shapes and a desired number of shape elements (N, usually 100). The output template is concise (7N values) and can be rendered or converted to a mesh using techniques such as raytracing or marching cubes~\cite{lorensen1987marchingcubes}. 
\section{Related Work}

There is a long history of work on shape analysis aimed at
extracting templates or abstract structural representations for classes of shapes 
\cite{hu2016siggraph,hu2018functionality,laga20183d,mitra2014structure,xu2017data}.

\vspace*{2mm}\noindent{\bf Primitive Fitting:} Fitting of basic primitives is perhaps the oldest topic in 3D computer vision, beginning with Roberts~\cite{roberts1963} and continuing to today \cite{biederman1987recognition,kaiser2018survey,li2011globfit,li2018supervised}.
These methods focus on explaining individual observations with primitives, and do not necessarily provide consistency across different input shapes, so they cannot be used for the correspondence, transfer, and exploration applications
targeted in this paper.

\vspace*{2mm}\noindent{\bf Part Segmentation:}  Others have studied how to decompose mesh
collections into consistent sets of semantic parts, either through geometric
\cite{golovinskiy2009consistent} or learned methods
\cite{araslanov2016efficient,fish2014meta,kalogerakis2010learning,leon2017semantic}.  
These methods differ from ours in that
they depend on labeled examples to learn the shapes and arrangements of
{\em semantic parts} within specific classes.  In contrast, we aim to learn a 
{\em structural} template shape for any class without human input.

\vspace*{2mm}\noindent{\bf Template Fitting:}  The most related techniques to ours are methods that explicitly fit templates to shapes 
\cite{brunelli2009template}.  The templates can be
provided by a person~\cite{ganapathi2018parsing,ovsjanikov2011exploration}, derived from part segmentations~\cite{zheng2014recurring, li2017grass, fish2014meta}, or  
learned automatically~\cite{kim2013learning,tulsiani2017learning,yumer2012co,zheng2014recurring}.
Previous work generally assumes an initial set of primitives or part structure is given prior to learning. For example, Kim et al. \cite{kim2013learning} proposed an optimization to fit an initial set of box-shaped primitives to a class of 3D shapes and used them for correspondence and segmentation.  Part structure is assumed by \cite{zheng2014recurring}.

Others have learned shape templates with a neural network. In Zou et al.~\cite{zou20173d}, a supervised RNN is trained to generate sets of primitives matching those produced by a heuristic fitting optimization. Sharma et al.~\cite{sharma2018csgnet} use reinforcement learning to decompose input shapes into a CSG parse tree. Like our approach, this approach does not require additional training data, but CSG trees are unsuitable for many template applications.

Tulsiani et al. \cite{tulsiani2017learning}
proposed a neural network that learned placements for a small number (3 to 6) of
box primitives from image or shape inputs, without additional supervision.   
Our method builds on this approach, but greatly expands the number and detail of the shape elements, allowing for the precise shape associations required
for correspondence and semantic segmentation applications.

\vspace*{2mm}\noindent{\bf Implicit Shape Representations:}  Decades ago, research\-ers in computer graphics proposed representing shapes with sets of local shape functions~\cite{ricci1973constructive,blinn1982generalization}.
The most common form is a summation of polynomial or Gaussian basis functions centered at arbitrary 3D positions, sometimes called \emph{metaballs}~\cite{blinn1982generalization}, blobby models~\cite{muraki1991volumetric}, or soft objects~\cite{wyvill1986soft}.
Other forms include convolution surfaces \cite{bloomenthal1991convolution} and partition of unity implicits \cite{ohtake2003multi}.
These representations support compact storage, efficient interior queries, arbitrary topology, and smooth blends between related shapes, properties that are particularly useful for our application of predicting template shapes.

\vspace*{2mm}\noindent{\bf Shape Representations for Learning:}  Recently, several deep network architectures have appeared that encode observations (color images, depth images, 3D shapes, etc.) into a latent vector space and decode latent vectors to 3D shapes.  Our work follows this approach. We argue that our structured implicit representation is superior for template learning compared to decoding voxels \cite{brock2016generative,wu2016learning,wu20153d}, sparse-voxel octrees \cite{tatarchenko2017octree}, points~\cite{fan2017point}, meshes~\cite{groueix2018papier,kanazawa2018learning,wang2018pixel2mesh}, box primitives~\cite{tulsiani2017learning}, signed-distance function estimators~\cite{DeepSDF}, or indicator function estimators~\cite{mescheder2019occ}. 

Table~\ref{tab:representation_properties} compares the properties of these representations. Compared to points, implicit surfaces are superior because they provide a clearly-defined surface. Compared to meshes, implicit surfaces can continuously adapt to arbitrary topology. Structured implicit functions are most similar to voxel grids since both implicitly represent a surface. Unlike voxel grids, they provide a sparse representation of shape, though octree techniques can provide sparse representations of voxels. The major difference for our work is that our shape elements can be moved and transformed in a smooth way to, for example, track gradual changes in airplane wing shape across a shape collection. By contrast, two similar, but slightly transformed shapes will have entirely different voxel representations. 
\definecolor{Gray}{gray}{0.92}
\begin{table}
\centering
\begin{small}
\begin{tabular}{l|ccccccc|}
\hline
Property            & Voxel & Octree & Point   & Mesh   & Deep  & Ours \\
\hline
Interpret           & +    & +    & +   & +    & -    & + \\
\rowcolor{Gray}
Concise             & -    & +    & +    & +     & -    & + \\
Surface             & +     & +     & -    & +    & -     & +  \\
\rowcolor{Gray}
Volume              & +    & +    & -    & -     & +    & +  \\
Topology            & +    & +    & -    & -    & +   & + \\
\rowcolor{Gray}
Deform              & -    & -    & +    & +    & -    & + \\
\hline
\end{tabular}
\end{small}
\caption{Comparison of desirable properties of various 3D representations, rated as suitable (+) or unsuitable (-). From top to bottom: is the representation interpretable to humans; concise in storage; capable of representing surfaces and volumes; allows topological changes; and supports smooth deformation. Structured implicit functions are suitable in all properties. ``Deep'' refers to methods that represent a volumetric function as a deep neural network~\cite{DeepSDF,sitzmann2018deepvoxels}.}
\label{tab:representation_properties}
\end{table}

Techniques have recently been proposed to directly approximate volumetric functions such as signed-distance fields or indicator functions using deep neural networks~\cite{DeepSDF, mescheder2019occ,sitzmann2018deepvoxels}. Compared to these approaches, structured implicit functions are light weight, easily interpretable, and provide template geometry that can be modified or transformed by later processing.

\section{Structured Implicit Shape Representation}

\newcommand*{\pos}{\ensuremath{\mathbf{x}}}
\newcommand*{\tpar}{\ensuremath{\mathbf{\Theta}}}
\newcommand*{\epar}{\ensuremath{\mathbf{\theta}_i}}
\newcommand*{\simp}{\ensuremath{F(\pos, \tpar)}}

We assume each input shape can be modeled as a watertight surface bounding an interior volume (real-world meshes usually must be processed to satisfy this assumption, see Sec.~\ref{sec:mesh_conversion}). We aim to represent this surface as the $\ell$ level set of a function $\simp$, where $\pos$ is a 3D position and $\tpar$ is a vector of template parameters. In the structured implicit formulation, $F$ is the sum of the contributions of a fixed number of shape elements with local influence, labeled $i\in [N]$, where $N$ is their count.  Each element is a function $f_i$ defined by its parameter vector $\epar$ (making $\tpar$ simply the concatenation of $\epar$):
\begin{equation}
    \simp = \sum_{i \in [N]} f_i(\pos, \epar)
\end{equation}

The specific version of shape elements we adopt are \emph{scaled axis-aligned anisotropic 3D Gaussians}. Here, $\epar$ consists of a scale constant $c_i$, a geometric center $\mathbf{p}_i \in \mathcal{R}^3$, and per-axis radii $\mathbf{r}_i \in \mathcal{R}^3$.
\begin{equation}
    f_i(\pos, \epar) = c_i \exp\left(\sum_{d \in \{x,y,z\}} \dfrac{-(\mathbf{p}_{i,d} - \pos_d)^2} {2\mathbf{r}_{i,d}^2}\right)
\end{equation}

Intuitively, one can think this of representation as a set of squished or stretched 3D blobs. We found this set of parameters to be the minimum necessary to achieve good results. More sophisticated shape elements, such as full multivariate Gaussians, or even windowed quadric functions~\cite{ohtake2003multi}, would likely improve results, but we do not experiment with those here.  

Because all constants $c_i$ are negative, we have that $f_i(\pos, \epar) < 0$ and thus $\simp < 0$, $\forall \pos \in \mathcal{R}^3$. Therefore we pick a negative isolevel $\ell$ and define the surface $S$ to be its crossing:
\begin{equation}
    S = \left\{ \pos  \in \mathcal{R}^3 : \simp  = \ell\right\}
\end{equation}

We set $\ell := -0.07$, which was chosen by grid search. The reason that the constants are negative rather than positive is to maintain the convention that function values inside the surface should be less than $\ell$, while values outside the surface should be greater than $\ell$. This leads to a convenient binary outside/inside test for points $x$:

\begin{equation}
    \simp > \ell
\end{equation}

For most experiments presented here, we use $N = 100$. Because each shape element has seven parameters, the total dimensionality of our representation is a fixed $7N = 700$ floating point values.

\section{Template Learning}
\begin{figure*}
    \centering
    \includegraphics[width=\textwidth]{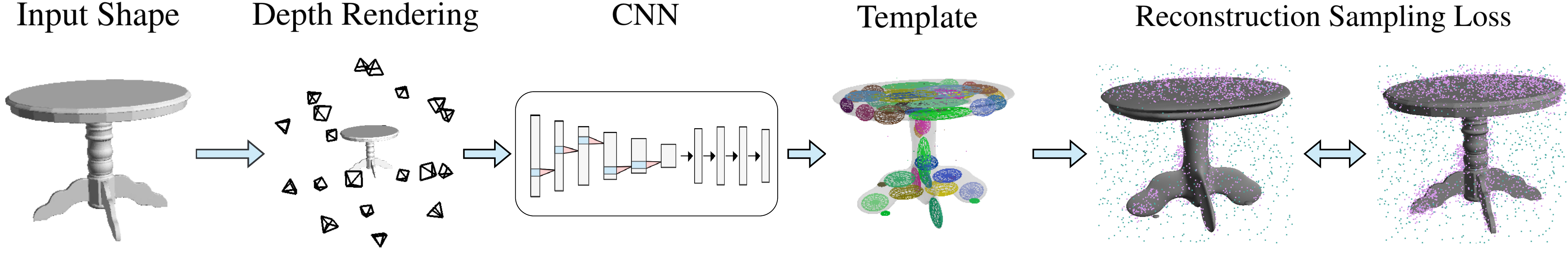}
    \caption{An overview of our method. The input to our system is a mesh. We render a stack of depth images around the mesh, and provide these as input to an early-fusion CNN.  The output of the CNN is a vector with fixed dimensionality. This vector is interpreted as a shape template with parameters that define an implicit surface. Next, we sample points near the ground truth surface and also uniformly in space. A classification loss enforces that each sample point is correctly labeled as inside/outside by the surface reconstruction. }
    \label{fig:fig_method}
\end{figure*}
We propose a learning framework (Figure~\ref{fig:fig_method}) to train a neural network to fit the shape template to data. The network's goal is to find the template parameters $\tpar$ that best fit a 3D shape, where the loss penalizes the amount of predicted shape that is on the wrong side of the ground truth inside/outside border. 
We render multiple depth images of the mesh from fixed views to provide 3D input to the network. Our network has a feed-forward CNN architecture and predicts the entire parameter vector $\tpar$ at once with a fully connected layer. During training, we choose sparse sample locations in 3D and evaluate our loss function at those locations with a classification loss. The details of this procedure are described in the rest of this section.

Note that although fitting consistency is vital to our applications, we do not directly enforce similar shapes to have similar template parameters; the network arrives at a smooth fitting function without intervention. We hypothesize that, as a matter of optimization, the smooth solution is ``easier'' for the network to learn, but analyzing the causes of this behavior is an engaging direction for future work.

\subsection{Architecture}
In order to learn the template, we first need to encode the input 3D shape. There are a variety of network architectures for encoding 3D shape; options include point networks~\cite{qi2016pointnet}, voxel encoders~\cite{maturana2015voxnet}, or multi-view networks~\cite{su15mvcnn}. Because voxel encoders can be computationally expensive, and point cloud encoders discard surface information, we opt for a multi-view encoding network. Our network takes a stack of 20 depth images rendered from the vertices of a dodecahedron as input, as in~\cite{kanezaki2018rotationnet}. The
network contains 5 convolutional layers followed by 4 fully connected layers.

The final fully connected layer is linear and maps to the template parameter vector $\tpar$, which in our experiments is usually 700-D. 
Even though we use an encoder/decoder style architecture, there is no heavy decoding stage: the code vector is our explicit representation. We experimented with alternative ``decoding'' architectures, such as an LSTM that predicts each shape element in succession. We found the LSTM architecture to perform better in some cases, but it took much longer to train, and was not able to scale easily to large numbers of shape elements.

\subsection{Data Preparation}
\label{sec:mesh_conversion}
Before training, we must preprocess the input meshes to make them watertight. This step is important primarily because our loss function requires a ground truth inside/outside classification label.  

In order to do the watertight conversion, we first convert the meshes to a $300^3$ sparse voxel representation~\cite{museth2013vdb}. We flood fill the octree to determine inside/outside, then extract the isocontour of the volume to produce the watertight mesh. We generate 100,000 random samples uniformly in the bounding box of the mesh, and compute 0/1 inside/outside labels. We additionally compute 100,000 samples evenly distributed on the surface of the mesh.

We also render depth maps of the watertight meshes. For each mesh, we render 20 depth images at uniformly sampled viewing directions as input to the network. The (depth maps, labeled samples) pairs are the only data used for learning.

\subsection{Loss}
The goal of our loss function is only to measure deviation from the input shape; we assume that our representation will naturally create a smooth template due to its structure. In order to accurately reconstruct the surface, we employ three individual loss functions, described in detail in the following sections. $L_U$ and $L_S$ are classification losses ensuring that the volume around the ground truth shape is correctly classified as inside/outside. These losses were inspired by recent work on implicit function learning~\cite{mescheder2019occ, chen2018learning}. $L_C$ enforces that all of the shape elements contribute to the reconstruction. The total loss function is a weighted combination of the three losses:
\begin{equation}
    L = w_U L_U + w_S L_S + L_C
\end{equation}
$L_C$ has no weight here because it contains two subclases with different weights $w_a$ and $w_b$.

As our losses compare the structured implicit value $\simp$ to indicator function labels (0 inside, 1 outside), we formulate a soft classification boundary function to better facilitate gradient learning:
\begin{equation}
    G(\pos, \tpar) = \mathrm{Sigmoid}\left(\alpha (\simp - \ell)\right)
\end{equation}
where $\alpha$ controls the sharpness of the boundary, and is set to 100 as determined by grid search.

\subsubsection{Uniform Sample Loss $L_U$}
If $\simp$ correctly classifies every point in the volume according to the ground truth shape boundary, then it has perfectly reconstructed the ground truth. To measure the classification accuracy, we choose $(x,y,z)$ coordinates uniformly at random in the bounding box of the ground truth mesh. We evaluate $\simp$ at these locations, and apply a loss between the softened classification decision $G$, and the ground truth class label, which is 0 inside and 1 outside:
\begin{equation}
    L_U(\pos, \tpar) = \begin{cases}
    \beta G(\pos, \tpar)^2 & \pos \ \mathrm{inside} \\
    (1 - G(\pos, \tpar))^2& \pos \ \mathrm{outside}\\
    \end{cases}
    \label{eq:bounding_box_loss}
\end{equation}

At each training batch we randomly select 3,000 of the precomputed 100,000 points to evaluate the loss. $\beta$ accounts for the inside/outside sample count differences.

\subsubsection{Near Surface Sample Loss $L_S$}
While the uniform sample loss is effective, it is problematic because it prioritizes surface reconstruction based on the fraction of the volume that is correct. The network can easily achieve 99\%+ correct volume samples and still not visually match the observation. In particular, thin structures are unimportant to a volumetric loss but subjectively important to the reconstruction. To improve performance, we sample proportionally to surface area, not volume. We additionally want to ensure that the network is not biased to produce an offset surface, so the loss should be applied with similar weight on both the positive and negative side of the surface boundary.

In order to achieve these goals, we implemented the following algorithm. 
For each of the 100,000 surface samples, a ray is cast in each of the positive and negative normal directions away from the surface point. Because the mesh is watertight, at least one of the two samples must intersect the surface. The minimum of these two intersection distances is chosen, and truncated to some threshold. We sample a point along either normal direction with probability inversely proportional to the squared distance from the surface and proportional to the minimum intersection distance. The output samples roughly satisfy both of our goals: no thin structures are missed, regardless of their volume, and there is an equal sampling density on both sides of the surface. 

This loss function, $L_S$, is identical to $L_U$ (see Equation~\ref{eq:bounding_box_loss}) except for the sample locations where it is applied. Note that $L_S$ and $L_U$ are not redundant with one another. Because $L_S$ only contains samples very near the surface, it does not on its own enforce that the network keep free space clear of spurious shapes. 
We found it most effective to use a weighted combination of both losses, using $L_S$ to do hard example mining, and $L_U$ to ensure that free space around the shape remains clear.

\subsubsection{Shape Element Center Losses $L_C$}
One problem with the loss so far is that it is only concerned with the final composite function $\simp$. If shape elements do not affect $F$, they also don't affect the loss. This ``death'' of shape elements can easily happen over time, since elements are randomly initialized and some are likely to be far from the ground truth surface. Their contribution to $L_U$ and $L_S$ is small, and there is no incentive for the network to use them. Our solution to this problem is to apply a third loss $L_C$, the center classification loss. This loss enforces that all predicted centers must lie on the inside of the predicted shape and within the ground truth bounding box:
\begin{equation}
    L_C(\pos, \tpar) = \begin{cases}
    w_a G(\pos, \tpar)^2 & \pos \in B\\
    w_b \sum_d \max(0, B_L - \pos_d, B_U - \pos_d)^2 & \pos \notin B
    \end{cases}
\end{equation}

Above, $w_a$ and $w_b$ are hyperparameters balancing the two cases, which are in different units. $B$ is the axis aligned bounding box of the ground truth shape, which has a lower coordinate $B_L$ and an upper coordinate $B_U$. It states that if the predicted center $\pos$ is inside the ground truth bounding volume (where $L_U$ will be applied, keeping free space empty), then $\pos$ must also be inside the predicted surface. On the other hand, if $\pos$ is outside the ground truth bounding boxing, then it should be directly encouraged to move inside the bounding volume because it can't be useful to the template from that distance.

\section{Experiments}

We conduct experiments to demonstrate important properties of the shape template: it accurately fits a wide variety of shapes, fits similar shapes with similar templates, can be used to find 3D-to-3D and 2D-to-3D correspondences, and can be fit from RGB images alone. We train and test on ShapeNet Core V2~\cite{chang2015shapenet}, using the dataset split defined by 3D-R2N2~\cite{choy20163d}. We show results trained on both the full dataset (Sections~\ref{sec:clustering}, \ref{sec:single_view}, \ref{sec:shape_correspondence}) and trained per-class (Sections~\ref{sec:volumetric_primitives} and \ref{sec:shape_interpolation}). Identical hyperparameters we used to train all templates.

\subsection{Clustering by Template Parameters}
\label{sec:clustering}
\begin{figure}[t]
    \centering
    \includegraphics[width=0.9\columnwidth]{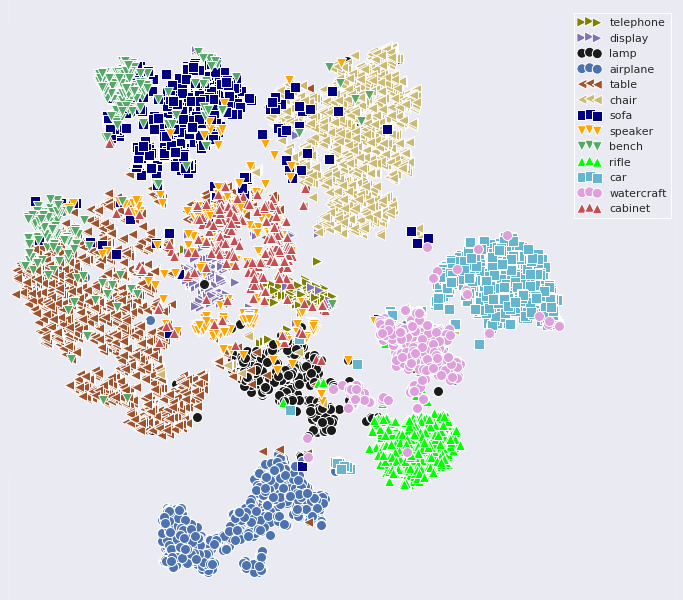}
    \caption{t-SNE visualization of template parameters on ShapeNet test set, colored by shape class labels. Note the clean clustering of most classes. Mixed clusters are also intuitive, e.g. mixing between tables, benches, and sofas.}
    \label{fig:multiclass_tsne}
\end{figure}
A desirable property of a template fitting procedure is that similar shapes are fit with similar template parameters. Figure~\ref{fig:multiclass_tsne} shows a t-SNE~\cite{vanDerMaaten2008tsne} visualization of the template parameter vectors $\tpar$ for the ShapeNet test set, colored by ShapeNet class labels. Several classes of shapes (airplanes, rifles, cars) are neatly clustered by their template parameters. Other classes are mixed, but in intuitive ways: some benches look like tables, other benches look like sofas, and some sofas look like chairs. Cabinets, speakers, and displays are all essentially boxes, so they have similar template parameters. 

\subsection{Comparison to Volumetric Primitives}
\label{sec:volumetric_primitives}
\begin{figure}[t]
    \centering
    \begin{tabular}{c}
    \includegraphics[width=\columnwidth]{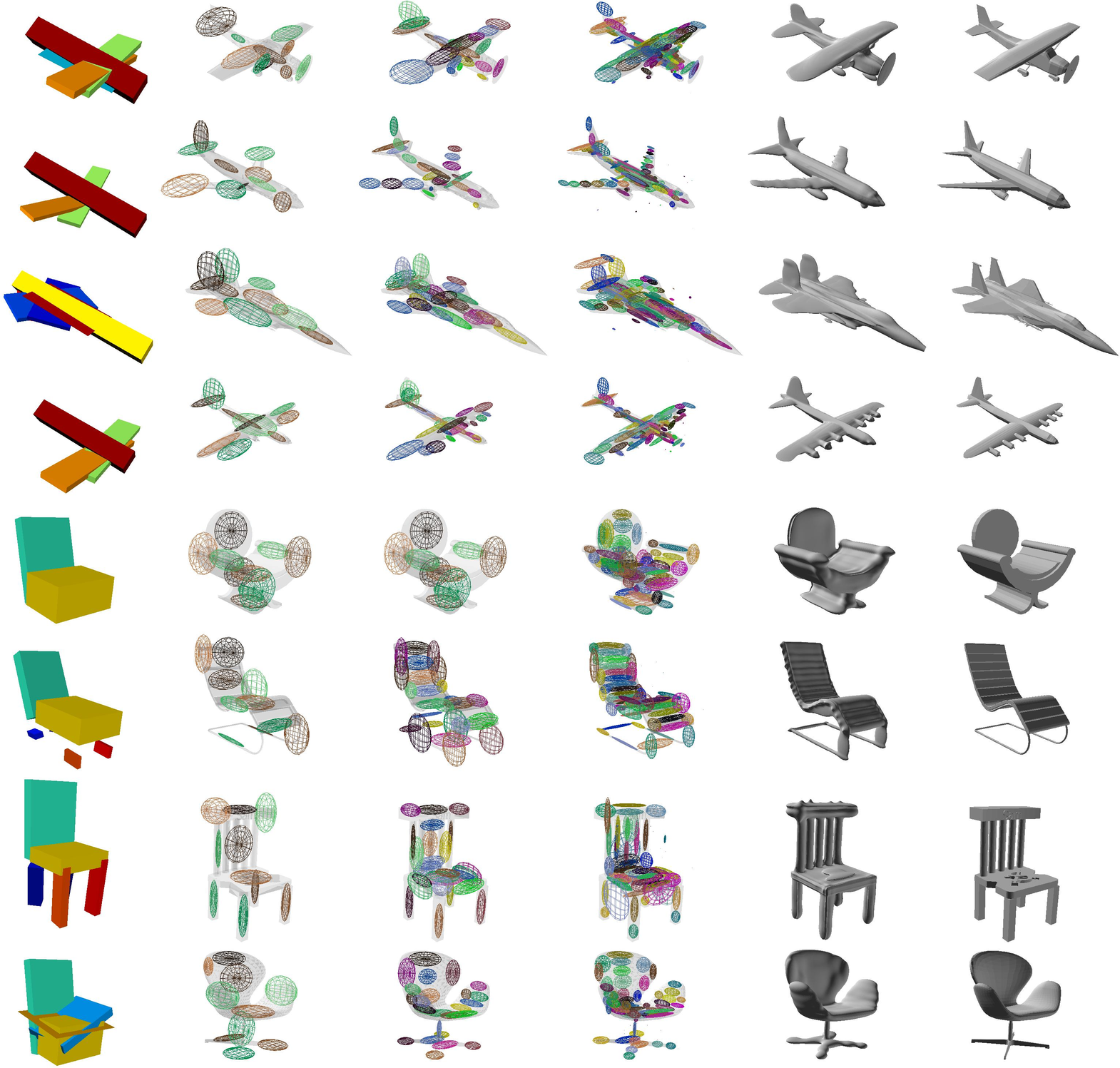}\\
    \hspace{-1em}\rule{0.95\columnwidth}{0.5pt}\\
    \includegraphics[width=\columnwidth]{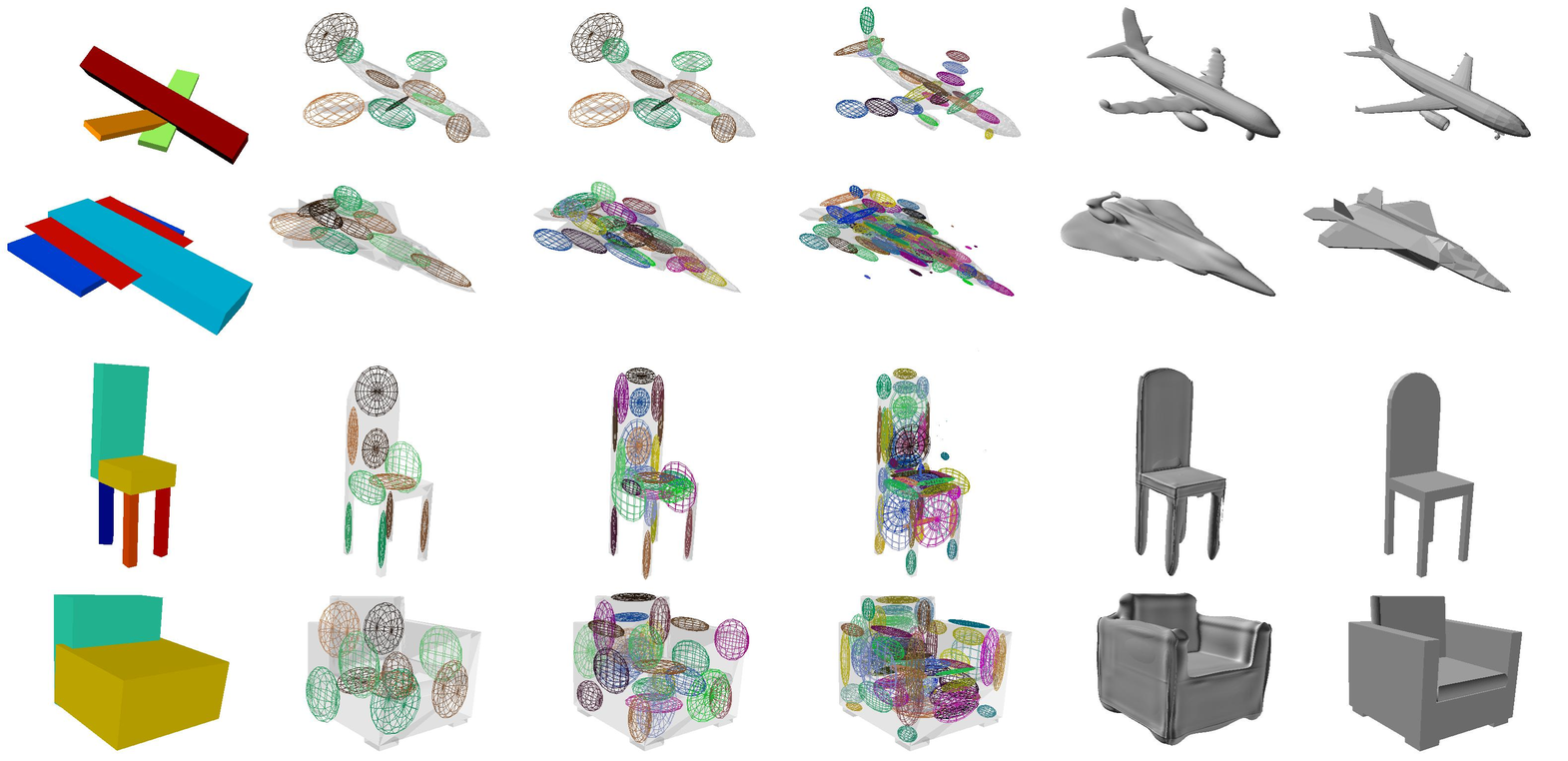}
    \end{tabular}
    \begin{scriptsize}
    \begin{tabular}{p{1.05cm}p{1.05cm}p{1.05cm}p{1.05cm}p{1.05cm}p{1.05cm}}
    a) VP~\cite{tulsiani2017learning} & b) N=10 & c) N=25 & d) N=100 & e) Recon & f) GT 
    \end{tabular}
    \end{scriptsize}
    \caption{Comparison to Volumetric Primitives~\cite{tulsiani2017learning}: (a) volumetric primitives result; (b-d) templates computed with our method for 10, 25, and 100 elements; (e) surface reconstruction from the template in (d); (f) ground truth surface mesh.  Shapes above the line come from our training set, while the shapes below the line are from our test set.}  
    \label{fig:tulsiani_comparison}
\end{figure}

The closest alternative approach to ours is the volumetric primitives of Tulsiani, et al.~\cite{tulsiani2017learning}. We provide a detailed comparison between our template shapes and their shape abstractions using results generously provided by the authors. For this comparison we trained one fitting network per shape class, not one network for all classes, to match the procedure of \cite{tulsiani2017learning}. Figure~\ref{fig:tulsiani_comparison} shows representative results for examples from the ShapeNet training set, with 10, 25, and 100 shape elements (see supplemental material for the full set of results). 
In comparison to volumetric primitives (Figure~\ref{fig:tulsiani_comparison} a), our templates (b-d) are more detailed, have higher consistency, and better reflect the structure of the input mesh (f).

\subsection{Single-View RGB Prediction and Labeling}
\label{sec:single_view}
\begin{figure}[t]
    \centering
    \includegraphics[width=\columnwidth]{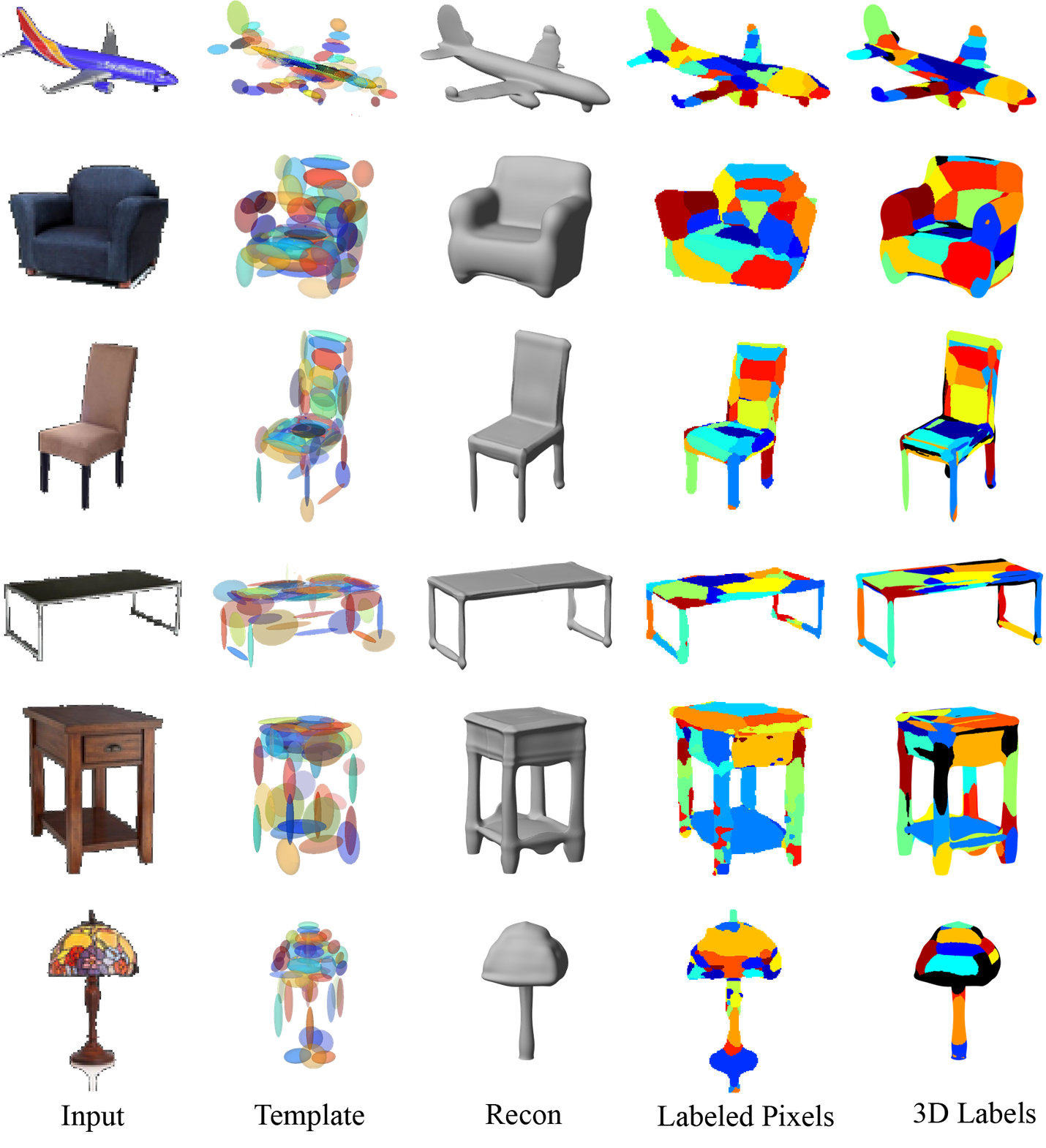}
    \caption{Template fitting and labeling from photographs. From left to right: input image with background removed, fit template,  corresponding isosurface, image pixels labeled by the highest-value shape element, corresponding 3D regions labeled by the highest-value element. Regions in 3D not found by the image labeling network are black. The labeling performs well for easily oriented shapes (top rows), and worse for shapes with rotational symmetries (bottom rows). Note that the labeling is based entirely on the template, without additional region or part labels.}
    \label{fig:2d3d_correspondence}
\end{figure}
Figure~\ref{fig:2d3d_correspondence} shows qualitative results demonstrating predictions from photographs of ShapeNet-style objects. To predict the template parameters from an RGB image, we apply a similar technique to CNN purification~\cite{li2015jointembeddings} or network distillation~\cite{hinton2015distilling} and train a second network that regresses from RGB to the template parameters already found through our 3D-to-3D training scheme. The training data for this network is synthetic OpenGL renderings of the ShapeNet training set, with camera angles chosen randomly from a band around the equator of the shape.

Because the template is consistent, we can go further than overall 3D shape prediction and predict correspondence between pixels in the image and the influence regions of individual shape elements (Figure~\ref{fig:2d3d_correspondence}, right). Each element tends to produce a particular part of each shape: the $i^{th}$ element might produce the tail fin of an airplane, while the $j^{th}$ might produce the wingtip. Because of this consistency, a semantic segmentation network~\cite{ronneberger2015unet} can be trained to label pixels by the index of the shape element with maximum weight at that pixel. The result is a segmentation of the image into 3D regions, without additional region or part labels. One limitation of this approach is that the template learning does not take into account object symmetry. Shapes with natural orientations, such as airplanes and chairs, are successful, while shapes without fronts and backs, such as the lamp and nightstand, confuse the network.

Similar techniques have been used for human body pose prediction~\cite{wei2016dense, bogo2016smplify} using hand-made templates, but to our knowledge, we are the first to use a learned template.

\subsection{Shape Correspondence}
\label{sec:shape_correspondence}
\begin{figure}[t]
    \centering
    \includegraphics[width=\columnwidth]{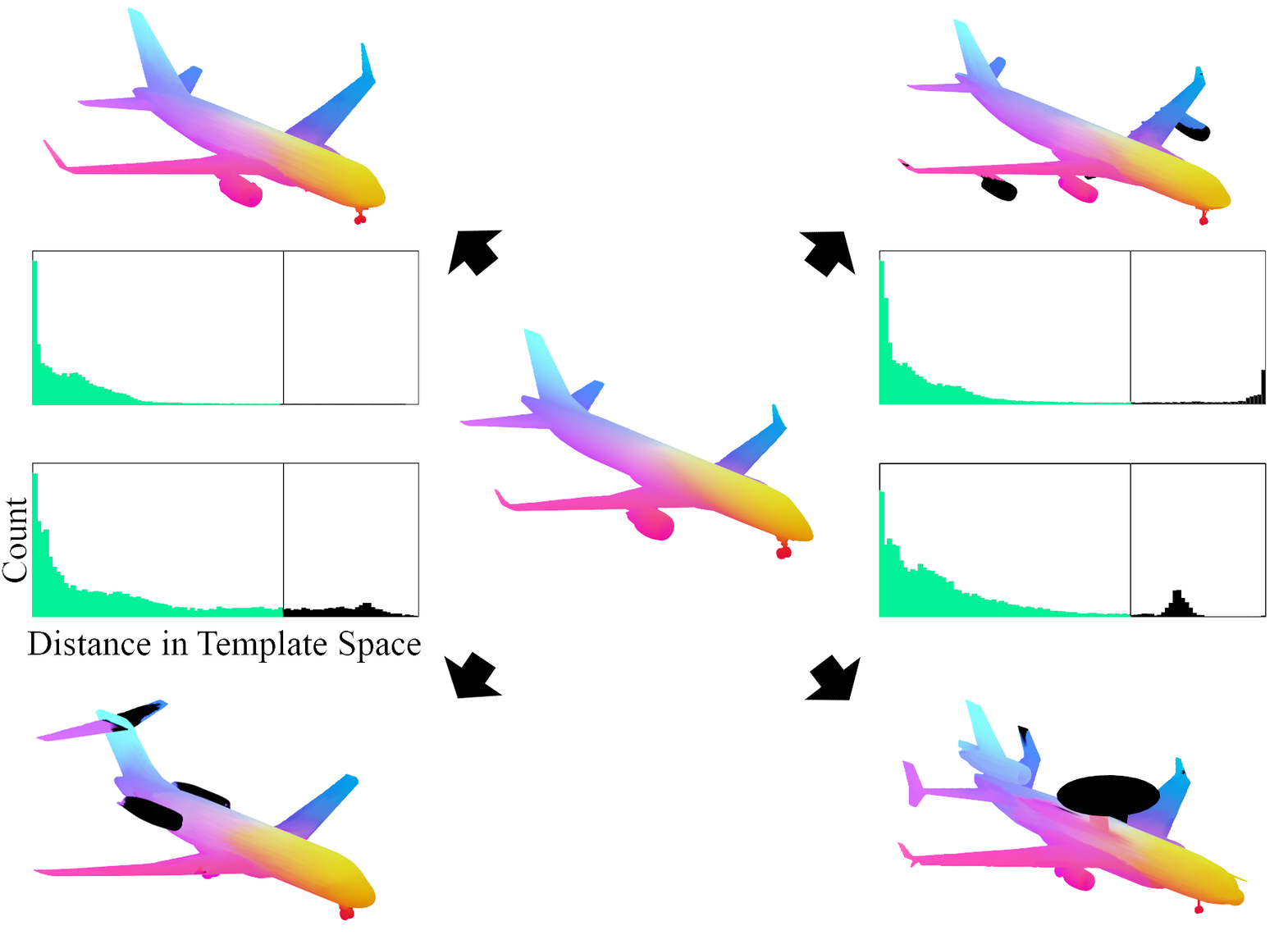}
    \caption{Transferring per-vertex colors from source airplane (center) to target airplanes (corners). Vertices are corresponded to their nearest neighbor in template space. Matching colors indicate corresponding vertices, while black regions have no corresponding vertices in the source.  The histograms plot the proportion of nearest-neighbor distances that produce good matches (green) and outliers (black, distance $>$ 0.65). Outliers include extra wing and tail engines, landing gear, and a radar dome, all missing on the source airplane.  Correspondences were computed for resampled ShapeNet meshes from the training set of the multi-class network.}
    \label{fig:3d3d_correspondence}
\end{figure}
The learned template is consistent across shapes of the same class, meaning that the same elements will influence equivalent shape parts (e.g. airplane wings).  This property can be exploited to find correspondences between different shapes.  We present one automatic approach to achieve that.  First, we use our network to compute the template configuration $\tpar$ of each shape we want to correspond.  Then, for each vertex $\mathbf{v}$, we compute its template coordinates.  The template coordinates consist of three numbers for each shape element.  Those are computed by subtracting the element's center from the vertex position, dividing each coordinate by the corresponding element radius (improving correspondence between elongated and squashed elements), then scaling that vector to be of length $F(\mathbf{v}, \tpar)$.  The direction of each per-element vector helps geometrically localize the vertex, while its length denotes the influence of that element.  Finally, the cosine distance between template coordinates can be used to find the closest target vertex for each source vertex, as visualized in Figure~\ref{fig:3d3d_correspondence}.

\subsection{Shape Interpolation}
\label{sec:shape_interpolation}
\begin{figure}[t]
    \centering
    \includegraphics[width=0.9\columnwidth]{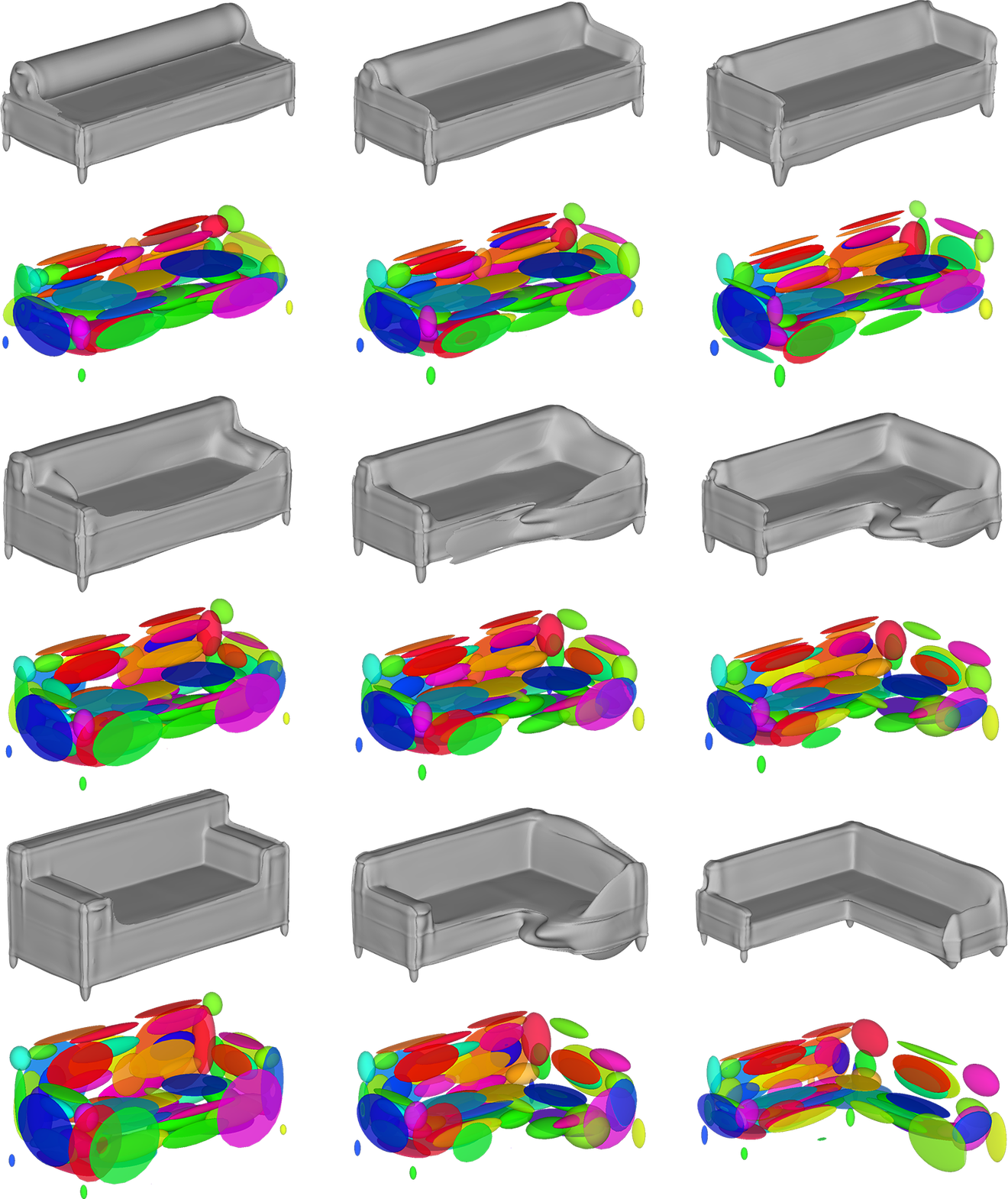}
    \caption{Linear interpolation between shape templates and corresponding surfaces of four sofas.  The reconstructed shapes are at the four corners, pairwise blends are in-between the neighboring corners, and the average of all four shapes is in the center.  The blends were obtained from training data of a single-class network.}
    \label{fig:lerp}
\end{figure}
Another benefit of the structured template is the ability to geometrically interpolate between shapes.  One can simply take multiple templates, blend all their parameters with per-template weights, and obtain an in-between template and hence the in-between shape.  Figure~\ref{fig:lerp} shows interpolation between four sofas computed this way.

\subsection{RGB Single View 3D Reconstruction}
\setlength{\tabcolsep}{2pt}
\begin{table}[t]
\centering
\begin{small}
\begin{tabular}{lcccc|cccc}
\toprule
Threshold  & \multicolumn{4}{c|}{$\tau$} & \multicolumn{4}{c}{$2\tau$} \\
\cmidrule(lr){2-5} \cmidrule(lr){6-9}
Category   & R2N2 & PSG & P2M & Our       & R2N2 & PSG & P2M & Our   \\
\midrule
plane      & 41 & 68 & \textbf{71} & 69    & 63 & 81 & 81 & \textbf{86} \\
bench      & 34 & 49 & 58 & \textbf{62}    & 49 & 69 & 72 & \textbf{82} \\
cabinet    & 50 & 40 & \textbf{60} & 40    & 65 & 67 & \textbf{77} & 64  \\
car        & 38 & 51 & \textbf{68} & 47    & 55 & 78 & \textbf{84} & 70  \\
chair      & 40 & 42 & \textbf{54} & 40    & 55 & 64 & \textbf{70} & 64  \\
monitor    & 34 & 40 & \textbf{51} & 42    & 48 & 64 & \textbf{67} & 65  \\
lamp       & 32 & 41 & \textbf{48} & 32    & 44 & 59 & \textbf{62} & 52  \\
speaker    & 45 & 32 & \textbf{49} & 29    & 58 & 57 & \textbf{66} & 50  \\
firearm    & 28 & 70 & \textbf{73} & 72    & 47 & 83 & 83 & \textbf{88}  \\
sofa       & 40 & 37 & \textbf{52} & 42    & 53 & 63 & \textbf{70} & \textbf{70}  \\
table      & 44 & 53 & \textbf{66} & 40    & 59 & 73 & \textbf{79} & 61  \\
cellphone  & 42 & 56 & \textbf{70} & 56    & 61 & 80 & \textbf{83} & 79  \\
watercraft & 37 & 51 & \textbf{55} & 49    & 52 & 71 & 70 & \textbf{75}  \\
\midrule
mean       & 39 & 49 & \textbf{60} & 48    & 55 & 70 & \textbf{74} & 70  \\
\bottomrule
\end{tabular}
\end{small}
\caption{F-score (\%) on ShapeNet test set, with $\tau = 10^{-4}$ as in \cite{wang2018pixel2mesh}. Higher numbers are better. }
\label{tab:reconstruction_results}
\end{table}
While exact shape reconstruction is not the focus of our work, we compared the reconstruction accuracy of the template surface with the output of 3D-R2N2~\cite{choy20163d}, Point Set Generation Network~\cite{fan2017point}, and Pix2Mesh~\cite{wang2018pixel2mesh}. The inputs are single RGB images of unknown camera orientation, so we use the distillation approach from Section~\ref{sec:single_view}. The train/test split is from 3D-R2N2. Our shape representation has only 700 degrees of freedom, compared with $32^3 = 32768$ DoF for the 3D-R2N2 voxel grid, $1024*3 = 3072$ DoF for points generated by PSG, and $2466*3 = 7398$ DoF for the vertices of the Pix2Mesh mesh. Despite having many fewer degrees of freedom, the template surface reconstruction accuracy is similar to competing approaches (Table~\ref{tab:reconstruction_results}).

\subsection{Limitations}
\begin{figure}[t]
    \centering
    \begin{tabular}{ccccc}
    \includegraphics[height=1.1in]{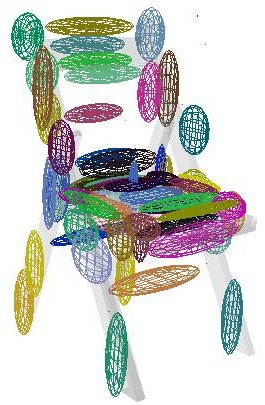} &
    &
    \includegraphics[height=1.1in]{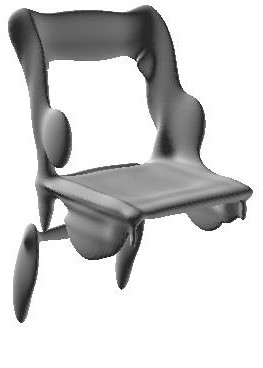} &
    &
    \includegraphics[height=1.1in]{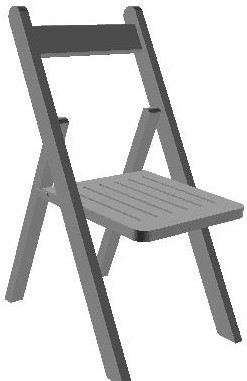} \\
    a) Template fit & & b) Reconstruction & & c) Input mesh
    \end{tabular}
    \caption{Shapes with angled parts, sharp creases, 
    and thin structures are difficult for our method to learn.}  
    \label{fig:limitations}
\end{figure}

Our method has several limitations apparent in Figure \ref{fig:limitations}, which exhibits several failure cases. First, since our representation comprises of a small number of axis-aligned functions, it has limited ability to represent detailed, sharp, or angled structures (e.g., creases or corners).  Second, since it learns to classify sides of a surface boundary, it struggles to reconstruct razor thin structures.  Finally, since it uses a fixed number of shape elements (e.g., 100), it does not produce a template with 1-to-1 mapping to semantic shape components.  We believe these limitations could be addressed with alternative (higher-order, non axis-aligned) local functions, distance-based loss functions, supervised training, and/or network architecture search.

\section{Conclusion}
This paper investigates using structured implicit functions to learn a template for a diverse collection of 3D shapes.   We find that an encoder-decoder network trained to
generate shape elements learns a template that maps detailed surface geometry consistently across related shapes in a collection with large shape variations.  Applications for the learned template include shape clustering, exploration, abstraction, correspondence, interpolation, and image segmentation.  Topics for future work include learning to generate higher-order and/or learned shape elements, deriving semantically meaningful shape elements via supervised learning, and using structured implicit functions for other applications such as 3D reconstruction.

\section{Acknowledgements}
We acknowledge ShapeNet~\cite{chang2015shapenet}, 3D-R2N2~\cite{choy20163d}, and the Stanford Online Products Dataset~\cite{song2016deep} for providing training and evaluation data for our method. We also thank the authors of Volumetric Primitives~\cite{tulsiani2017learning} for providing extended results from their method for our comparisons. We thank Avneesh Sud for helpful discussions and comments.

{\small
\bibliographystyle{style/ieee}
\bibliography{bibliography}
}

\clearpage
\appendix

\section{Additional Algorithmic Details}

\vspace*{2mm}\noindent{\bf Hyperparameters:} We first detail the hyperparameters associated with training our 3D $\rightarrow$ template network in Table~\ref{tab:depth_hparams}. The goal of our architecture design was to enable encoding 3D shape into our representation in a robust and generalizable way. As a result, the architecture is a sequence of convolutional layers followed by a sequence of fully connected layers, and was trained using Adam~\cite{kingma15adam}. We experimented with more complex architectures, such as ResNet~\cite{he2016resnet}, but found them to be slow to converge and to generalize poorly (likely due to lack of sufficient training data to train an architecture like ResNet from scratch).

\begin{table}[b]
\centering
\begin{tabular}{||c|c||}
    \hline
    Name & Value\\
    \hline
    $\alpha$ & 100.0 \\
    $\beta$ & 10.0\\
    $w_U$ & 1.0\\
    $w_S$ & 0.1\\
    $w_a$ & $10 / 3$\\
    $w_b$ & 0.01\\
    $\ell$ & -0.07\\
    Batch Size & 8\\
    Learning Rate & $5\times10^{-5}$\\
    Adam $\beta_1$ & 0.9\\
    Adam $\beta_2$ & 0.999\\
    Uniform Samples & 3000\\
    Near Surface Samples & 3000\\
    Conv. Width(s) & 3\\
    Conv. Stride(s) & 2\\
    Conv. Depths & 16, 32, 64, 128, 128\\
    FC Widths & 1024, 2048, 2048, 7*N\\
    Nonlinearity & Leaky ReLU\\
    Input Resolution & 137x137x20\\
    \hline
\end{tabular}
\vspace{1em}
\caption{Hyperparameters and optimization details for the early fusion depth image network. This network is composed of 5 convolutional layers followed by 4 fully connected layers.}
\label{tab:depth_hparams}
\end{table}

\vspace*{2mm}\noindent{\bf Single View Reconstruction Architecture:}  We next describe the architecture used for our RGB $\rightarrow$ template experiments (Table~\ref{tab:rgb_hparams}).  In this application, we use ResNet V2 50 to encode images.  We find that using this network was more effective than the architecture used for encoding depth images, likely for at least two reasons. First, it could leverage initial weights pretrained on ImageNet classification~\cite{russakovsky15imagenet} (all layers except for the final two FC layers were pretrained). Second, it had larger capacity to utilize the larger RGB dataset -- while there were approximately 30,000 shapes in the 3DR2-N2~\cite{choy20163d} training split of ShapeNet~\cite{chang2015shapenet}, there are over 700,000 images in the 3DR2-N2 RGB render training split. This is because there are 24 RGB renders of each ShapeNet shape in the 3DR2-N2 set. This acts as a form of data augmentation, which can be leveraged by a larger network.

\begin{table}[t]
    \centering
    \begin{tabular}{||c|c||}
    \hline
        Name & Value\\
        \hline
        Batch Size & 128\\
        Learning Rate & $5\times 10^{-5}$ \\
        Backbone Network & ResNet V2 50\\
        Pretraining & ImageNet\\
        Finetuning & End-to-end\\
        FC Widths & 2048, 7*N\\
        Image Native Resolution & 137x137 (3D-R2N2)\\
        Network Input Resolution & 224x224\\
        Adam $\beta_1$ & 0.9\\
        Adam $\beta_2$ & 0.999\\
    \hline
    \end{tabular}
    \vspace{1em}
    \caption{Hyperparameters and optimization details for the RGB $\rightarrow$ 3D network. This network is composed of ResNet V2 50, up to and including the average pooling layer, followed by two fully connected layers that map to the 3D representation.}
    \label{tab:rgb_hparams}
\end{table}

Note that when training our RGB network, we used network distillation~\cite{hinton2015distilling} rather than the loss functions described in the main paper. In particular, the only loss for the RGB network was a supervised $L^2$ regression loss to the ``ground truth'' template generated by our depth network for each training example.

\section{Additional Dataset Details}
For all experiments, we used the Shapenet~\cite{chang2015shapenet} 80\%-20\% train-test split provided by 3D-R2N2~\cite{choy20163d}. Because this split does not include a validation set, we further randomly split train into a 75\%-5\% train-val split. All networks were trained on train only (not train+val). Val was used to choose hyperparameters, although some experiments in the paper are primarily concerned with train performance.

In addition to the 3D-R2N2 training set, we generated and trained on a larger, more diverse set of 2 million RGB renders of ShapeNet. The renders have a higher native resolution (256x256, compared to 137x137). We also added additional data augmentation during training, varying the brightness, contrast, hue, and saturation randomly. We qualitatively found this to significantly improve domain transfer from synthetic renders to real images. However, for consistency, all results reported for the RGB $\rightarrow$ template network were trained on 3D-R2N2 renders. For the semantic segmentation network, which was never tested on 3D-R2N2, all results were trained on the larger dataset.

\begin{table}[h]
    \centering
    \begin{tabular}{||c|c||}
    \hline
        Name & Value\\
        \hline
        Batch Size & 8\\
        Learning Rate & $1\times 10^{-4}$ \\
        Encoder Network & ResNet V2 50\\
        Decoder Network & U-Net based on \cite{ronneberger2015unet}\\
        Pretraining & None\\
        Input Resolution & 256x256\\
        Output Resolution & 256x256\\
        Adam $\beta_1$ & 0.9\\
        Adam $\beta_2$ & 0.999\\
    \hline
    \end{tabular}
    \vspace{0.5cm}
    \caption{Hyperparameters for the 2D semantic segmentation model.}
    \label{tab:segm_hparams}
\end{table}
\section{Runtime Analysis}

\vspace*{2mm}\noindent{\bf Inference time:} Because our depth $\rightarrow$ template network is a small feed-forward CNN, inference for a template given depth renderings of a mesh is very fast. Using a GTX 1080 with a batch size of 1, mean network inference time is 1.11ms. 

\vspace*{2mm}\noindent{\bf Training time:} Since inference is so quick, the performance bottleneck shifts to pre/post processing, such as rendering depth images of the input mesh. The time required to do this is highly dependent on the complexity of the input mesh (although for most of ShapeNet the rasterization process was bottlenecked by the cost of writing the output images to disk). 

\vspace*{2mm}\noindent{\bf Surface reconstruction:} Though not the goal of our work, structured implicit functions provide a direct way to extract a surface reconstruction (isosurface) from template parameters after inference.   Several methods are available with different accuracy vs. time trade-offs.  

One possibility is to render the isosurface of the representation directly using ray marching. One benefit of this approach is that the resulting image will accurately reflect the predicted isosurface and its normals. However, achieving interactive framerates with this approach is difficult.

A second alternative is to using marching cubes to extract the isosurface as a mesh. This has the advantage of quick rasterization and it makes comparisons to ground truth meshes easier, so it is the technique we apply in this paper.
For our surface extraction implementation, the primary computational bottleneck is to sample $F(\mathbf{x}, \mathbf{\Theta})$ at the marching cubes locations.  The simplest implementation would iterate over the elements of the marching cubes grid and compute $F(\mathbf{x}, \mathbf{\Theta})$ directly at each location. However, it is much more efficient to take advantage of the local nature of $f_i(\mathbf{x}, \theta_i)$. In particular, we set a minimum ``influence'' epsilon for $\frac{f_i(\mathbf{x}, \theta_i)}{c_i}$. This function is the component of $f_i(\mathbf{x}, \theta_i)$ that falls off with distance and is always in $[0, 1]$. In practice we set $\epsilon = 10^{-3}$. Then, we iterate over each shape element, and add its marginal contribution to the marching cubes volume by iterating only over the voxels where it has nontrivial contribution. With this algorithm, mesh extraction takes approximately 5.96 seconds at $256^3$ on the CPU with one thread. GPU acceleration for this task is also possible, which could result in further runtime decreases.

We also postprocess the extracted isosurface by removing connected components below a trivial size threshold. This is helpful because one common artifact is for a function $f_i(\mathbf{x}, \theta_i)$ to be almost off, but still cross $\ell$ within a small region of space. As the marching cubes resolution increases, smaller and small volumes can be revealed, so this post processing step is useful. A surface area threshold of $0.005$ for connected components was used for the interpolation video.

\end{document}